\documentclass{article} 
\usepackage{iclr2027_conference,times}

\usepackage{amsmath,amsfonts,bm}

\usepackage[utf8]{inputenc} 
\usepackage[T1]{fontenc}    
\usepackage{hyperref}       
\usepackage{url}            
\usepackage{booktabs}       
\usepackage{nicefrac}       
\usepackage{microtype}      
\usepackage{wrapfig}
\usepackage{enumitem}
\usepackage{CJKutf8}
\usepackage{makecell}
\usepackage{array}
\usepackage{capt-of}
\usepackage{xcolor}
\usepackage{mdframed}
\usepackage{tabularx}
\usepackage{colortbl}
\usepackage{amssymb}
\usepackage{subfigure}
\usepackage{multirow}
\usepackage{pifont}
\usepackage{arydshln}
\usepackage{fontawesome5}
\usepackage{tcolorbox}
\tcbuselibrary{breakable,skins}
\usepackage{algorithm}
\usepackage{algpseudocode}

\usepackage[table]{xcolor} 
\usepackage{ulem} 
\usepackage{tikz} 

\newmdenv[
  topline=false,
  bottomline=false,
  rightline=false,
  leftline=true,
  linecolor=black!35,
  linewidth=0.7pt,
  backgroundcolor=white,
  innerleftmargin=8pt,
  innerrightmargin=0pt,
  innertopmargin=4pt,
  innerbottommargin=4pt,
  skipabove=6pt,
  skipbelow=6pt
]{taskvariant}

\definecolor{mlb}{RGB}{173,216,230}  
\definecolor{mlo}{RGB}{255,223,186}  

\def\eqref#1{equation~\ref{#1}}

\def\1{\bm{1}}

\DeclareMathAlphabet{\mathsfit}{\encodingdefault}{\sfdefault}{m}{sl}
\SetMathAlphabet{\mathsfit}{bold}{\encodingdefault}{\sfdefault}{bx}{n}

\usepackage{hyperref}
\usepackage{url}

\title{SWE-MILE: Asynchronous Potential-Induced Milestone Credit Assignment for Long-Horizon Software Engineering Agents}

\author{
Chaoqun Cui$^{1,2}$,~Hao Zhou$^3$,~Meiqi Chen$^3$,~Fandong Meng$^{3,*}$,~\textbf{Wenji Mao}$^{1,2,*}$\\
$^1$MAIS, Institute of Automation, Chinese Academy of Sciences\\
$^2$School of Artificial Intelligence, University of Chinese Academy of Sciences\\
$^3$Tencent Inc., China\\
\texttt{cuichaoqun2025@ia.ac.cn, wenji.mao@ia.ac.cn}
}

\iclrfinalcopy 
\definecolor{varblue}{RGB}{0,120,215}
\newtcolorbox{promptbox}[1]{
    title=#1,
    colback=gray!5!white,
    colframe=gray!70!black,
    coltitle=white,
    breakable,
    fontupper=\small
}

\begin{document}

\maketitle
\pagestyle{plain}

\begin{abstract}

Long-horizon software engineering (SWE) agents trained with reinforcement learning with verifiable rewards (RLVR) typically receive only terminal outcome supervision, making it difficult to distinguish productive actions from redundant exploration or functional regressions. We propose SWE-MILE, an asynchronous potential-induced milestone credit assignment framework that derives fine-grained process supervision from workflow runtime, without auxiliary reward models or external evaluators. SWE-MILE quantifies task-relevant file exposure and test-state alignment as navigation and verification potentials, respectively. Differences in these potentials attribute milestone progress and regressions to individual actions, while discounted backward credit propagates supervision to preceding steps. To efficiently acquire intermediate verification states, SWE-MILE further introduces asynchronous shadow probing, which replays repository-changing actions in an isolated sandbox and runs verification in parallel with the agent's primary interaction, largely hiding verification latency. The resulting process credit augments terminal outcome advantages and provides informative learning signals. Experiments on two representative long-horizon SWE tasks demonstrate substantial improvements in agent performance, highlighting workflow runtime signals as a practical source of process supervision for long-horizon SWE agents.

\end{abstract}

\section{Introduction}

In recent years, Large Language Models (LLMs) have evolved from isolated code generation toward long-horizon software engineering (SWE) agents for repository-level tasks \citep{swebench,sweagent,openhands}. These SWE agents operate in real code repositories through multi-turn interaction, performing codebase navigation, inspection, editing, and testing to accomplish issue resolution and even whole-repository generation tasks \citep{swebench,nl2repobench}. With the development of large-scale executable SWE environments such as SWE-Gym and R2E-Gym \citep{swegym,r2egym}, reinforcement learning with verifiable rewards (RLVR) has gradually become an important training paradigm for further improving the ability of SWE agents to autonomously solve complex tasks \citep{longcontextswe,gra}. These methods use unit-test-based task-specific verifiers provided by SWE environments to provide objective reward feedback for agent policy rollouts, and then optimize the agent policy with GRPO, PPO, and their variants \citep{deepswe,grpo,ppo}. However, this reliable feedback is still predominantly consumed as a terminal outcome. The final verifier determines whether a rollout succeeds, but does not reveal which intermediate actions contribute to or undermine that outcome.

RLVR relying solely on such coarse-grained supervision faces a severe temporal credit assignment problem. SWE trajectories typically contain tens or hundreds of heterogeneous actions. A successful trajectory may contain redundant exploration or temporary functional regressions, while a failed trajectory may still correctly localize the issue or partially fix it. Therefore, broadcasting the same outcome reward to all actions in a trajectory makes it difficult to reflect each action's contribution to the final outcome. Existing work explores fine-grained supervision from three directions: 1. \textit{Rule-based methods}, which assign process rewards based on predefined interaction rules, such as valid tool calls, successful tool execution, or other execution validity signals \citep{gra,contextfolding,reponavigator}. However, these signals mainly capture protocol compliance or shallow tool behaviors rather than substantive task progress or functional regressions, while handcrafted rules remain susceptible to reward hacking. 2. \textit{Auxiliary model-based methods}, which evaluate intermediate actions by training reward models or prompting external LLM evaluators with process-specific rubrics \citep{sweshepherd,swetrace,rubricgrm}. These methods rely on model-estimated proxy signals that are not directly grounded in verifiable functional changes of the repository, while introducing additional inference workload and latency. 3. \textit{State aggregation methods}, which aggregate steps reaching equivalent states across multiple rollouts and estimate step values from their downstream outcomes \citep{gigpo,rtmc,graphgpo}. However, the rich environment states in SWE tasks make state matching sparse, while process values are still inferred indirectly from downstream outcomes.

In this study, we explore a complementary perspective: the runtime of a long-horizon SWE workflow itself provides a rich yet underutilized source of process supervision. Although the final verifier outcome is binary, its underlying verification evidence is inherently structured and fine-grained. For example, a bug repair task requires a set of Fail-to-Pass (F2P) and Pass-to-Pass (P2P) tests to reach their target states. While the final success/failure is a binary reduction, the runtime exposes much richer states, such as which target tests have already passed or which previously correct functionalities have regressed. Beyond verifier-derived evidence, the agent's continuous interaction with the environment also reveals other critical progress signals, such as whether key code files have been retrieved and inspected. The key challenge is how to attribute these milestone progresses to specific steps, so as to assign quantitative process rewards according to the progress or regression caused by the agent's actions.

Specifically, to address how to capture milestone progress signals naturally produced during workflow runtime and attribute these signals to specific actions as process rewards, we propose SWE-MILE, an asynchronous potential-induced milestone credit assignment framework for agentic RL across long-horizon SWE tasks, including issue resolution and whole-repository generation. SWE-MILE has the following properties: First, SWE-MILE relies on neither additional reward models nor LLM evaluators, and constructs process supervision solely from reliable intermediate information naturally available in the task data and environment. Second, SWE-MILE uses an asynchronous shadow sandbox to replay actions that modify the repository state and probe intermediate verification states, without introducing additional rollout latency. Third, SWE-MILE organizes the exposure of task-relevant files in the agent context and the alignment between verification states and target states into a quantifiable potential, and assigns rewards to specific actions according to increases and regressions in this potential. This provides effective process signals and helps prevent reward hacking by discouraging behaviors such as deliberately breaking and then repairing functionality or repeatedly inspecting the same file.

In summary, the main contributions of this study are as follows.
\begin{itemize}
\item We introduce runtime-grounded process supervision, a process reward paradigm for long-horizon SWE agent RLVR that derives action-level credit from observable task progress naturally produced during workflow runtime.
\item We propose SWE-MILE, which quantifies the exploration coverage of task-relevant context as navigation potential and measures verification potential through low-latency asynchronous shadow probing, deriving fine-grained process credit from their potential differences.
\item Experimental results show that agent policies trained with SWE-MILE achieve substantial performance improvements on both issue resolution and whole-repository generation tasks.
\end{itemize}

\section{Method}

\subsection{Preliminaries}

\noindent\textbf{Problem Formulation.} Given an SWE task $q$, the agent policy $\pi _{\theta }$ interacts with the environment $\mathbb{E}$ over multiple steps to complete the task. At step $i$, $\pi _{\theta }$ generates reasoning $r_i$ and an executable action $a_i$ based on the current environment state $s_{i}$. After executing the action in $\mathbb{E}$, the environment state transitions to $s_{i+1}=\mathbb{E}(s_i,a_i)$. This process forms a trajectory $\mathcal{T}$:
\begin{equation}
\mathcal{T}=\{q,s_1,r_1,a_1,s_2,\cdots ,s_{n-1},r_{n-1},a_{n-1},s_n\}.
\end{equation}
At the final step, the verifier of the SWE task, typically a task-specific test suite, assigns a binary outcome reward $R$ to the terminal environment state $s_n$. $R$ cannot distinguish the actual contributions of different actions to task completion. Therefore, the goal of step-level credit assignment is to assign each step a step-level reward, which should reflect the progress or regression caused by the agent output $(r_i,a_i)$.

\noindent\textbf{Terminology.} We define three terms used throughout this study. A \textit{milestone} refers to observable intermediate progress or regression along a trajectory. A \textit{potential} is a scalar state function that quantifies such progress. A \textit{shadow probe} refers to running the verifier once in an isolated shadow sandbox to obtain the intermediate test states of the repository at a given step.

\subsection{Overall Framework}

\begin{figure*}[t]
  \centering
  \includegraphics[width=\textwidth]{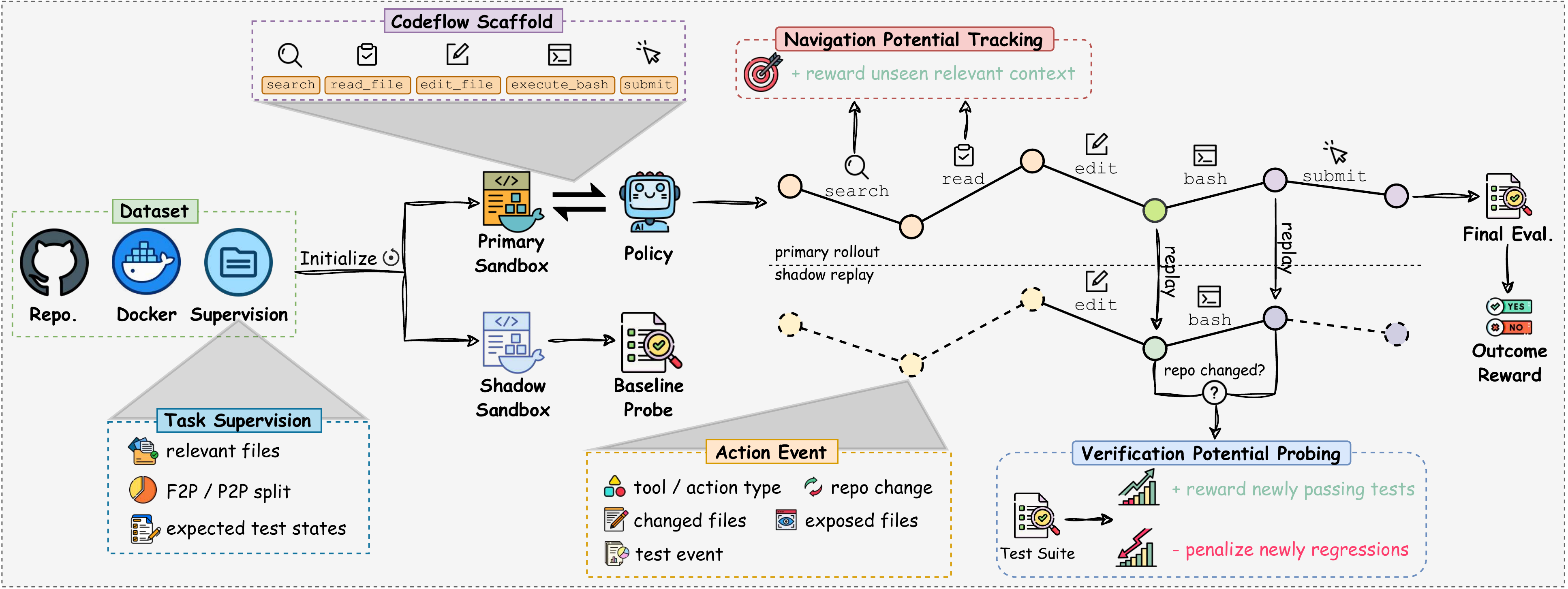}
  \caption{Overview of SWE-MILE. Navigation potential tracks the exposure of task-relevant context during primary interaction, while repository-changing actions are asynchronously replayed in a shadow sandbox to probe verification potential and provide fine-grained process rewards.}
  \label{fig:framework}
\end{figure*}

As shown in Figure~\ref{fig:framework}, SWE-MILE extracts verifiable milestone credit from SWE agent interaction trajectories while retaining the terminal outcome reward. SWE-MILE defines the coverage of task-relevant files in the agent context as \textit{navigation potential}, and the alignment of the current verification states of tests with their target states as \textit{verification potential}. Progress or regression is then attributed to specific actions through potential differences between adjacent steps, while discounted backward credit propagates weaker rewards or penalties to preceding actions that contribute to subsequent potential changes. For each rollout, a primary sandbox and a shadow sandbox are initialized. The agent interacts with the environment in the primary sandbox, while the shadow sandbox replays repository-changing actions in their original order and asynchronously probes intermediate verification states in parallel with the primary rollout. During training, this process credit is used to adjust the RLOO outcome advantage  \citep{rloo}, providing long-horizon SWE agents with fine-grained, high-confidence process supervision signals.

\subsection{Codeflow Interaction Harness}

We equip the agent with a scaffold containing multiple structured tools, termed \textit{Codeflow}, which separates retrieval, inspection, editing, and testing into individual actions to facilitate attribution of agent behavior. Specifically, the Codeflow scaffold provides five structured tools: \texttt{search}, \texttt{read\_file}, \texttt{edit\_file}, \texttt{execute\_bash}, and \texttt{submit}. SWE-MILE attributes navigation potential based on tool invocation behavior. Despite the availability of dedicated tools, the agent may still use \texttt{execute\_bash} to perform file retrieval, inspection, and editing. Codeflow therefore applies static Shell analysis based on lexical tokenization and command rules to identify retrieval and inspection operations, and combines the parsed operation semantics with the standard output actually visible to the model for file-level navigation attribution. Editing behavior is attributed based on repository hash consistency before and after tool execution. In addition, to study the impact of a structured action space versus a more flexible Bash-only action space on agent performance, Codeflow also supports a Bash only mode that provides only the \texttt{execute\_bash} and \texttt{submit} tools. Further details of the Codeflow scaffold are provided in Appendix~\ref{sec:expcodeflow}.

\subsection{Milestone Potential and Process Credit}

Inspired by \citet{potential}, we derive the process reward of SWE-MILE from milestone potential differences. We instantiate runtime-grounded process supervision with two milestone potentials, navigation and verification, as representative signals of intermediate progress. The same principle can extend to other readily observable, task-general runtime states. 

\noindent\textbf{Navigation Potential.} For an SWE task, such as an issue resolution task, navigation potential is defined as the coverage of relevant files that are effectively discovered and inspected by the agent. Supervision for the set of relevant files comes from annotations naturally produced during the construction of datasets such as R2E-Gym or from the set of files involved in the gold patch. Navigation potential measures the extent to which file information required for correct modifications has been explored during the rollout or exposed in the model context. Specifically, let the set of relevant files be $\mathcal{H}$. At step $t$, the cumulative exposure of each file $f\in \mathcal{H}$ is denoted by $q_{f,t}\in[0,1]$:
\begin{equation}
q_{f,t}=\max(q_{f,t-1},q_{f,t}^{\text{observed}}),
\end{equation}
where the larger value between the historical cumulative exposure and the observation at the current step is retained. At step $t$, if file $f$ does not appear, then $q_{f,t}^{\text{observed}}=0.0$; if its path appears in search results, then $q_{f,t}^{\text{observed}}=0.2$; if its content is successfully exposed to the agent, then $q_{f,t}^{\text{observed}}=1.0$. This design prevents repeated retrieval or inspection of the same relevant file from continuously increasing the navigation potential, thereby preventing reward hacking. Intuitively, once a file has been exposed in the agent context, repeated searches or inspections do not provide additional effective exposure. The navigation potential at step $t$ is defined as:
\begin{equation}
\Phi_t^{\text{navi}}=\frac{1}{|\mathcal{H}|}\sum_{f\in \mathcal{H}}q_{f,t}.
\end{equation}

Importantly, navigation potential must be scored based on the agent's actual observation rather than the command string alone. Failed commands should not receive credit; for example, using \texttt{read\_file} to inspect a file and receiving \texttt{Error: file not found} should not be counted. Some datasets may provide line-level annotations for relevant files, enabling more precise quantification of agent navigation behavior. In this study, however, we use only file-level supervision.

\noindent\textbf{Verification Potential.} The agent's objective is to align the verifier test states with their target states, rather than to produce the same solution as the gold patch. For issue resolution tasks with F2P/P2P test partitions, let $\mathcal{F}=\{i:y_{i,0}\ne y_i^*\}$ denote the set of F2P tests that have not yet reached their target states initially, and let $\mathcal{G}=\{j:y_{j,0}=y_j^*\}$ denote the set of tests that are initially correct and should remain so. At a repository-changing step $t$, the current test states $y_{\cdot,t}$ are obtained through a shadow probe. For repository-unchanged steps, the test states remain the same as in the preceding step. The fraction of F2P tests that have been fixed at the current step is: 
\begin{equation}
p_t=\frac{1}{|\mathcal{F}|}\sum_{i\in \mathcal{F}}\mathbf 1[y_{i,t}=y_i^*],
\end{equation}
and the fraction of P2P tests that have regressed is:
\begin{equation}
b_t=\frac{1}{|\mathcal{G}|}\sum_{j\in \mathcal{G}}\mathbf 1[y_{j,t}\ne y_j^*].
\end{equation}
The verification potential at step $t$ is then defined as:
\begin{equation}
\Phi_t^{\text{veri}}=p_t-\beta_{\text{any}}\mathbf 1[b_t>0]-\beta_{\text{frac}}b_t.
\end{equation}
where the hyperparameters can be set to $\beta_{\text{any}}=0.1$ and $\beta_{\text{frac}}=0.5$. $\beta_{\text{any}}$ ensures that any regression causes at least a certain decrease in potential, preventing the impact of a single regression from being excessively diluted when the number of P2P tests is large. $\beta_{\text{frac}}$ controls the penalty strength for regression severity. In the initial state, $p_0=0$, $b_0=0$, and thus $\Phi_0^{\text{veri}}=0$. At the target state, where all F2P tests are repaired with no regressions, $\Phi^{\text{veri}}=1$. When substantial regressions occur, $\Phi_t^{\text{veri}}$ may also become negative. This design accounts for the fact that issue resolution tasks typically contain one or even two orders of magnitude more P2P tests than F2P tests.

\begin{taskvariant}
For whole-repository generation tasks, the agent typically starts from an empty repository, while the verifier contains hundreds or even thousands of acceptance tests. In this setting, measuring verification potential using the normalized test pass rate would dilute the local credit provided by an individual test. Therefore, for whole-repository generation tasks, verification potential is measured by the unnormalized pass count difference from the initial baseline. Let the total number of acceptance tests be $N$, and let the number of tests matching their target states after step $t$ be:
\begin{equation}
M_t=\sum_{i=1}^{N}\mathbf 1[y_{i,t}=y_i^*],
\end{equation}
where the baseline number of passing tests $M_0$ is obtained through an initial shadow probe. The verification potential at step $t$ is then defined as:
\begin{equation}
\Phi_t^{\text{veri}}=M_t-M_0.
\end{equation}
Thus, the initial state satisfies $\Phi_0^{\text{veri}}=0$, while at the target state, where all acceptance tests pass, $\Phi^{\text{veri}}=N-M_0$.
\end{taskvariant}

\noindent\textbf{Process Reward and Credit.} For issue resolution tasks, the overall milestone potential at step $t$ is defined as the weighted sum of the navigation potential and verification potential:
\begin{equation}
\Phi_t=\alpha_{\text{navi}}\Phi_t^{\text{navi}}+\alpha_{\text{veri}}\Phi_t^{\text{veri}},
\end{equation}
where $\alpha_{\text{navi}}$ and $\alpha_{\text{veri}}$ control the relative strengths of the two signals and can be set to $\alpha_{\text{navi}}=0.05$, $\alpha_{\text{veri}}=0.2$. The process reward at step $t$ is defined as the potential difference between adjacent steps:
\begin{equation}
r_t^{\text{pot}}=\Phi_t-\Phi_{t-1}.
\end{equation}
This potential difference attributes milestone changes, such as newly exposed relevant files, repaired tests, or functional regressions, to the corresponding actions. Repeatedly inspecting files that have already been fully exposed yields no additional navigation reward, while deliberately breaking functionality and then repairing it is penalized before receiving reward for the subsequent recovery. Therefore, process rewards based on potential differences naturally prevent reward hacking.

\begin{taskvariant}
For whole-repository generation tasks, navigation potential is disabled. To limit the reward scale caused by large single-step changes in test outcomes, the verification potential difference is asymmetrically clipped before weighting:
\begin{equation}
r_t^{\text{pot}}=\alpha_{\text{veri}}\operatorname{clip}\!\left(\Phi_t^{\text{veri}}-\Phi_{t-1}^{\text{veri}},-c_{-},c_{+}\right),
\end{equation}
where $\alpha_{\text{veri}}=0.1$, $c_{-}=6$, $c_{+}=3$. This setting uses a smaller scaling factor for the verification potential based on the unnormalized pass count, while retaining a stronger penalty for regressions.
\end{taskvariant}

Next, we augment the process reward with backward credit and an output-format reward to obtain the process credit at step $t$:
\begin{equation}
C_t^{\text{process}}=r_t^{\text{pot}}+\lambda\sum_{u=t+1}^{|\mathcal{T}|}\gamma^{u-t}r_u^{\text{pot}}+\eta r_t^{\text{fmt}},
\end{equation}
where the hyperparameters are set to $\lambda = 0.2$, $\gamma = 0.9$, $\eta = 0.25$. Since achieving a milestone typically results from the cumulative progress of multiple preceding actions, for example, a correct edit often depends on preceding steps that correctly search for and localize the relevant code, we use this backward credit to provide delayed feedback to preceding actions. In addition, to ensure that the agent can produce valid tool call formats, we introduce a format reward, where $r_t^{\text{fmt}}=0$ for a correct output format and $r_t^{\text{fmt}}=-1$ otherwise.

\subsection{Policy Optimization}

Direct step-level normalization across long-horizon SWE trajectories suffers from trajectory length bias, coupling between outcome and process rewards through shared normalization statistics, and contextual misalignment across semantically incomparable steps \citep{drgrpo,pass,mogrpo,hgpo}. SWE-MILE therefore avoids step-level cross-trajectory normalization and instead incorporates process credits as a correction term to the outcome advantage. We adopt an RLOO-style \citep{rloo} outcome advantage, because its value always lies within $[-1,1]$, providing a clear scale reference for incorporating the process credit. For $K$ trajectories sampled for a task, let the terminal outcome reward of trajectory $k$ be $R_k\in\{0,1\}$. Using the mean outcome reward of the remaining trajectories as the leave-one-out baseline, the outcome advantage is computed as:
\begin{equation}
A_k^{\text{outcome}}=R_k-\frac{1}{K-1}\sum_{m\ne k}R_m.
\end{equation}
All steps within a trajectory share the same outcome advantage, i.e., $A_{k,t}^{\text{outcome}}=A_k^{\text{outcome}}$. The final advantage used for policy optimization at step $t$ is:
\begin{equation}
A_{k,t}=A_{k,t}^{\text{outcome}}+C_{k,t}^{\text{process}}.
\end{equation}
We then optimize the policy using a step-level PPO-style clipped loss:
\begin{equation}
\mathcal{L}(\theta)=-\mathbb{E}\!\left[\min\!\left(\frac{\pi_\theta(o_{k,t}\mid h_{k,t})}{\pi_{\mathrm{old}}(o_{k,t}\mid h_{k,t})}A_{k,t},\operatorname{clip}\!\left(\frac{\pi_\theta(o_{k,t}\mid h_{k,t})}{\pi_{\mathrm{old}}(o_{k,t}\mid h_{k,t})},1-\epsilon_{-},1+\epsilon_{+}\right)A_{k,t}\right)\right],
\end{equation}
where $o_{k,t}$ and $h_{k,t}$ denote the model output and context at step $t$ of trajectory $k$, respectively. This design preserves trajectory-level comparison through the outcome advantage while distinguishing the contributions of individual steps within the same trajectory through the process credit.

\subsection{Asynchronous Shadow Probing}
\label{sec:shadow_probe}

\begin{figure}[!h]
\centering

\begin{minipage}[t]{0.505\textwidth}
    \vspace{0pt}  
    \centering
    \includegraphics[width=\linewidth]{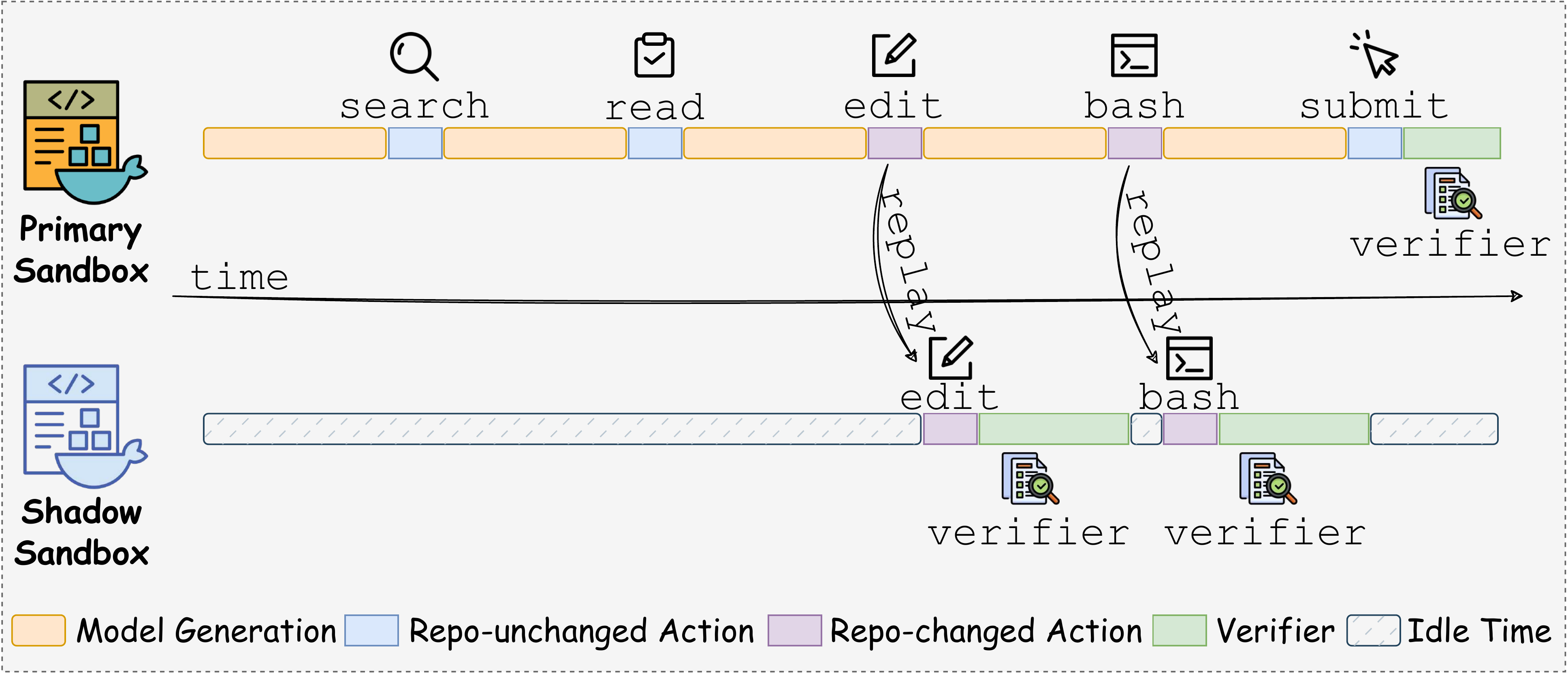}
    \vspace{-1.5em}
    \captionof{figure}{Asynchronous shadow probing timeline.}
    \label{fig:asynchronous}
\end{minipage}
\hfill
\begin{minipage}[t]{0.465\textwidth}
    \vspace{-0.9em}  
    \centering

    \captionof{table}{Runtime statistics of asynchronous shadow probing.}
    \label{tab:shadow_runtime}
    \vspace{0.8em}
    \resizebox{\linewidth}{!}{
    \begin{tabular}{lrrr}
        \toprule
        \textbf{Runtime Component} & \textbf{Mean} & \textbf{Unit} & \textbf{Samples} \\
        \midrule
        Agent generation - turn & 5.60 s & turn & 455,362 \\
        Agent generation - traj & 326.55 s & rollout & 9,600 \\
        \hdashline
        Action execution & 1.05 s & turn & 455,362 \\
        Shadow replay & 1.79 s & probe & 38,330 \\
        \hdashline
        Shadow verifier probe & 7.31 s & probe & 38,330 \\
        Primary final verifier & 8.90 s & rollout & 9,600 \\
        \hdashline
        Shadow tail latency & 2.07 s & rollout & 9,600 \\
        \bottomrule
    \end{tabular}
    }
\end{minipage}

\end{figure}

SWE-MILE creates a primary sandbox and a shadow sandbox with identical initial states for each rollout. The primary sandbox performs normal agent–environment interaction and compares the repository hashes before and after each action. When a hash change is detected, the corresponding action and its target repository state are asynchronously added to the shadow queue. The shadow sandbox replays these actions in trajectory order, runs the verifier test suite after each replay and obtains the intermediate test states used to compute the verification potential. Each such operation constitutes a shadow probe. As shown in Figure~\ref{fig:asynchronous}, because the shadow worker runs independently of the primary rollout, shadow probes typically overlap in time with the next round of model generation and subsequent interactions in the primary sandbox, requiring only a final synchronization at the end of the rollout. In Table~\ref{tab:shadow_runtime}, we report the runtime overhead across different stages for 9,600 rollouts from one SWE-MILE training run. The results show that verification potential probing introduces only about 2.07 s of wall-clock latency on average, indicating that sandbox parallelism allows the probing latency to be largely hidden by subsequent rounds of interaction.

During SWE agent RL training, both agent rollout and actor updates are GPU-bound computation, leaving substantial CPU and memory resources idle on training nodes \citep{dsec}. We therefore use the node-local SWE-MiniSandbox \citep{sweminisandbox} to deploy both the primary and shadow sandboxes. SWE-MiniSandbox maintains runtime isolation through namespaces and cgroups while sharing the underlying CPU and memory resources of the training node. This eliminates dependence on remote Docker sandbox services such as Kubernetes and their per-instance resource quotas, while improving resource utilization on local training machines.

\section{Experiments}

\subsection{Experimental Settings}

We evaluate SWE-MILE on issue resolution and whole-repository generation tasks. For issue resolution, we train on the Python subset (1,952 tasks) of the SWE-rebench V2-Filtered-Verified \citep{swerebenchv2} dataset, and evaluate on SWE-bench Verified (500 tasks) \citep{swebenchverified} and the public subset of SWE-bench Pro (731 tasks) \citep{swebenchpro}, with an inference budget of 200 turns and 256k context. For whole-repository generation, we train on the DeNovoSWE (3,675 tasks) \citep{denovoswe} dataset, and evaluate on NL2Repo-Bench (104 tasks) \citep{nl2repobench} and the Doc2Repo subset of BeyondSWE (50 tasks) \citep{beyondswe}, with an inference budget of 500 turns and 256k context. All training and evaluation experiments use the same Codeflow scaffold. More detailed experimental settings are provided in Appendix~\ref{sec:expdetailmain}.

Besides several frontier proprietary models, we compare against the following baselines:
\begin{itemize}
\item \textbf{GRPO} \citep{grpo} optimizes the policy using group-normalized advantages derived from outcome-level rewards.
\item \textbf{G-RA} \citep{gra} uses gated reward accumulation, incorporating rule-based immediate rewards for tool call formatting and selection preferences only when the terminal reward meets a predefined threshold.
\item \textbf{SWE-TRACE} \citep{swetrace} evaluates agent trajectories using task-specific rubrics and combines the resulting scores with terminal execution rewards to guide policy optimization.
\item \textbf{GiGPO} \citep{gigpo} augments trajectory-level group comparisons with step-level advantages by grouping actions that share the same state and comparing their subsequent returns.
\item \textbf{GraphGPO} \citep{graphgpo} merges sampled trajectories into a state-transition graph and assigns step-level credit according to how much each action reduces the shortest graph distance to a successful state.
\end{itemize}
Implementation details of these baseline methods are provided in Appendix~\ref{sec:baselineimp}. We additionally evaluate the DeNovoSWE-Agent-35A3B model \citep{denovoswe} on whole-repository generation tasks. The source code for SWE-MILE is available at \url{https://github.com/CcQunResearch/SWE-MILE}.

\subsection{Main Results}

In our experiments, we mainly compare different methods in terms of Avg@4 performance, average output token length, and average interaction turns.

\noindent\textbf{Issue resolution.} Table~\ref{tab:mainexp-issue} shows that on SWE-bench Verified and SWE-bench Pro, SWE-TRACE achieves strong task performance but requires more interactions, while GiGPO and GraphGPO better control trajectory length but still underperform SWE-MILE in task success. This suggests that the value of step-level feedback depends on how effectively it captures progress in repository repair. SWE-MILE quantifies state quality solely through potentials, whereas GiGPO and GraphGPO additionally require determining state equivalence. Given the rich and continuously changing state space of SWE tasks, even compressed state representations struggle to balance matching opportunities with progress information. This may explain why SWE-MILE improves Avg@4 while maintaining shorter responses and fewer interactions.

\begin{table}[!h]
    \centering
    \caption{Issue resolution results. Best and second-best results among process reward methods are in \textbf{bold} and \underline{underlined}, respectively.}
    \label{tab:mainexp-issue}

    \small
    \renewcommand{\arraystretch}{1.05}
    \setlength{\heavyrulewidth}{0.8pt}
    \setlength{\lightrulewidth}{0.4pt}

    \setlength{\tabcolsep}{6pt}

    \begin{tabularx}{\linewidth}{
        @{}
        >{\raggedright\arraybackslash}p{0.46\linewidth}
        *{2}{>{\centering\arraybackslash}X}
        @{}
    }
        \toprule
        \multicolumn{3}{c}{\textit{Proprietary Models}} \\
        \midrule

        \textbf{Models}
        & \textbf{SWE-bench Verified}
        & \textbf{SWE-bench Pro} \\
        \midrule

        GPT-5.4 / 5.5 / 5.6 Sol
        & -- / -- / --
        & 57.7 / 58.6 / 64.6 \\

        Claude Opus 4.6 / 4.8 / 5
        & 80.8 / 88.6 / 96.0
        & 53.4 / 69.2 / 79.2 \\

        Qwen-3.6-Max / 3.7-Max / 3.8-Max
        & -- / 80.4 / --
        & 57.3 / 60.6 / 67.7 \\

        DeepSeek V4-Flash / V4-Pro
        & 79.0 / 80.6
        & 52.6 / 55.4 \\

        GLM 4.6 / 5 / 5.2
        & 68.0 / 77.8 / --
        & -- / 55.1 / 62.1 \\

        \bottomrule
    \end{tabularx}

    \vspace{5pt}

    \setlength{\tabcolsep}{2.5pt}

    \begin{tabularx}{\linewidth}{
        @{}
        >{\raggedright\arraybackslash}p{0.16\linewidth}
        *{6}{>{\centering\arraybackslash}X}
        @{}
    }
        \toprule
        \multicolumn{7}{c}{\textit{Process Reward Methods}} \\
        \midrule

        \multirow{2}{*}{\textbf{Methods}} & \multicolumn{3}{c}{\textbf{SWE-bench Verified}} & \multicolumn{3}{c}{\textbf{SWE-bench Pro}} \\
        \cmidrule(lr){2-4}
        \cmidrule(lr){5-7}

        & \textbf{Avg@4$\uparrow$} & \textbf{\mbox{Resp. Length$\downarrow$}} & \textbf{Turns$\downarrow$} & \textbf{Avg@4$\uparrow$} & \textbf{\mbox{Resp. Length$\downarrow$}} & \textbf{Turns$\downarrow$} \\
        \midrule

        Base Model & 53.6 & 47426 & \textbf{79.6} & 30.8 & 32649 & \underline{85.8} \\
        GRPO & 58.2 & 29447 & 113.5 & 33.9 & 30054 & 111.6 \\
        G-RA & 57.9 & 28332 & 107.4 & 34.1 & 29325 & 114.1 \\
        SWE-TRACE & \underline{60.5} & 32332 & 117.3 & \underline{36.6} & 32452 & 120.6 \\
        GiGPO & 58.4 & \underline{27665} & 93.5 & 34.7 & 28112 & 91.2 \\
        GraphGPO & 59.9 & 27945 & 91.4 & 35.1 & \underline{27454} & 89.7 \\
        \midrule
        \textbf{SWE-MILE} & \textbf{63.8} & \textbf{25825} & \underline{81.8} & \textbf{39.3} & \textbf{25094} & \textbf{78.4} \\

        \bottomrule
    \end{tabularx}

\end{table}

\noindent\textbf{Whole-repository generation.} Table~\ref{tab:mainexp-repogene} shows that SWE-MILE achieves the highest Avg@4 on both benchmarks, with the shortest average response length and the fewest interaction turns. Although other methods improve performance, they do not consistently reduce the number of interaction turns. This suggests that, although whole-repository generation is a generation-intensive task, better performance does not necessarily require longer interaction trajectories. Assigning step-level credit based on intermediate verification progress helps the model focus on effective modifications earlier, thereby achieving better performance at lower interaction cost.

\begin{table}[!h]
    \centering
    \caption{Whole-repository generation results.}
    \label{tab:mainexp-repogene}

    \small
    \renewcommand{\arraystretch}{1.05}
    \setlength{\heavyrulewidth}{0.8pt}
    \setlength{\lightrulewidth}{0.4pt}

    \setlength{\tabcolsep}{6pt}

    \begin{tabularx}{\linewidth}{
        @{}
        >{\raggedright\arraybackslash}p{0.46\linewidth}
        *{2}{>{\centering\arraybackslash}X}
        @{}
    }
        \toprule
        \multicolumn{3}{c}{\textit{Proprietary Models}} \\
        \midrule

        \textbf{Models}
        & \textbf{Doc2Repo}
        & \textbf{NL2Repo-Bench} \\
        \midrule

        GPT-5.4 / 5.5 / 5.6 Sol
        & 61.64 / -- / --
        & 41.3 / 50.7 / 56.8 \\

        Claude Opus 4.6 / 4.8 / 5
        & 60.39 / -- / --
        & 49.8 / 69.7 / 75.3 \\

        Qwen-3.6-Max / 3.7-Max / 3.8-Max
        & -- / -- / --
        & 42.9 / 47.2 / 55.9 \\

        DeepSeek V4-Flash / V4-Pro
        & -- / 57.20
        & 39.4 / 38.5 \\

        GLM-4.6 / 5 / 5.2
        & -- / 56.76 / --
        & 17.5 / 35.9 / 48.9 \\

        \bottomrule
    \end{tabularx}

    \vspace{5pt}

    \setlength{\tabcolsep}{2.5pt}

    \begin{tabularx}{\linewidth}{
        @{}
        >{\raggedright\arraybackslash}p{0.187\linewidth}
        *{6}{>{\centering\arraybackslash}X}
        @{}
    }
        \toprule
        \multicolumn{7}{c}{\textit{Process Reward Methods}} \\
        \midrule

        \multirow{2}{*}{\textbf{Methods}} & \multicolumn{3}{c}{\textbf{Doc2Repo}} & \multicolumn{3}{c}{\textbf{NL2Repo-Bench}} \\
        \cmidrule(lr){2-4}
        \cmidrule(lr){5-7}

        & \textbf{Avg@4$\uparrow$} & \textbf{\mbox{Resp. Length$\downarrow$}} & \textbf{Turns$\downarrow$} & \textbf{Avg@4$\uparrow$} & \textbf{\mbox{Resp. Length$\downarrow$}} & \textbf{Turns$\downarrow$} \\
        \midrule

        Base Model & 48.4 & 36558 & \underline{78.7} & 25.7 & 68814 & \underline{107.3} \\
        GRPO & 51.3 & 34673 & 92.7 & 28.0 & 64322 & 124.3 \\
        SWE-TRACE & 51.9 & 33786 & 85.7 &\underline{30.9} & \underline{61245} & 114.6 \\
        GraphGPO & \underline{52.4} & \underline{33482} & 89.7 & 30.1 & 62113 & 112.7 \\
        DeNovoSWE-Agent & 52.2 & 57472 & 112.0 & 29.4 & 83128 & 156.6 \\
        \midrule
        \textbf{SWE-MILE} & \textbf{54.7} & \textbf{30473} & \textbf{72.6} & \textbf{33.1} & \textbf{56323} & \textbf{90.2} \\

        \bottomrule
    \end{tabularx}

\end{table}

\subsection{Ablation Study}

To examine the contribution of each component, we separately remove all milestone credit, navigation potential, verification potential, and backward credit, and replace the RLOO outcome advantage with the GRPO advantage. We also evaluate performance when only the Bash tool is provided. All experiments retain the output-format reward. Table~\ref{tab:ablation} shows that all ablations reduce Avg@4 on both benchmarks while increasing the average response length and interaction turns. Removing all milestone credit causes the largest degradation, indicating that terminal feedback alone is insufficient to effectively guide long-horizon repair. Among individual ablations, removing verification has a larger impact than removing navigation, suggesting that test state changes provide a stronger progress signal, while file localization remains complementary. The ablations of backward credit and RLOO advantage further demonstrate their effectiveness in SWE-MILE. When only the Bash tool is provided, Codeflow can identify search and inspection behavior only by parsing Bash commands, which reduces the recall of navigation potential attribution and leads to a slight performance drop.

\begin{table}[!h]
    \centering
    \caption{Ablation results on SWE-bench Verified and SWE-bench Pro.}
    \label{tab:ablation}

    \small
    \renewcommand{\arraystretch}{1.05}
    \setlength{\heavyrulewidth}{0.8pt}
    \setlength{\lightrulewidth}{0.4pt}
    \setlength{\tabcolsep}{2.5pt}

    \begin{tabularx}{\linewidth}{
        @{}
        >{\raggedright\arraybackslash}p{0.162\linewidth}
        *{6}{>{\centering\arraybackslash}X}
        @{}
    }
        \toprule
        \multirow{2}{*}{\textbf{Methods}} & \multicolumn{3}{c}{\textbf{SWE-bench Verified}} & \multicolumn{3}{c}{\textbf{SWE-bench Pro}} \\
        \cmidrule(lr){2-4}
        \cmidrule(lr){5-7}

        & \textbf{Avg@4$\uparrow$} & \textbf{\mbox{Resp. Length$\downarrow$}} & \textbf{Turns$\downarrow$} & \textbf{Avg@4$\uparrow$} & \textbf{\mbox{Resp. Length$\downarrow$}} & \textbf{Turns$\downarrow$} \\
        \midrule

        Base Model & 53.6 & 47426 & 79.6 & 30.8 & 32649 & 85.8 \\
        SWE-MILE & 63.8 & 25825 & 81.8 & 39.3 & 25094 & 78.4 \\
        \midrule
        w/o milestone & 59.4 & 29359 & 115.5 & 34.7 & 29156 & 115.6 \\
        \quad w/o navigation & 62.0 & 26319 & 87.3 & 38.0 & 26731 & 87.8 \\
        \quad w/o verification & 60.7 & 28489 & 96.4 & 35.8 & 27580 & 95.7 \\
        \quad w/o backward & 61.4 & 27332 & 94.4 & 38.3 & 27721 & 89.2 \\
        w/ GRPO adv. & 62.7 & 26997 & 89.2 & 38.2 & 26371 & 84.4 \\
        Bash only & 62.2 & 26212 & 85.2 & 38.6 & 25943 & 83.4 \\
        \bottomrule
    \end{tabularx}
\end{table}

In addition, Appendix~\ref{sec:training_dynamics} presents the training dynamics of multiple metrics under different potential settings for issue resolution, illustrating how model performance and behavior evolve throughout training.

\subsection{Hyperparameter Analysis}

\begin{figure*}[!h]
  \centering
  \subfigure[Navigation Potential Weight $\alpha_{\text{navi}}$]{\includegraphics[width=0.48\textwidth]{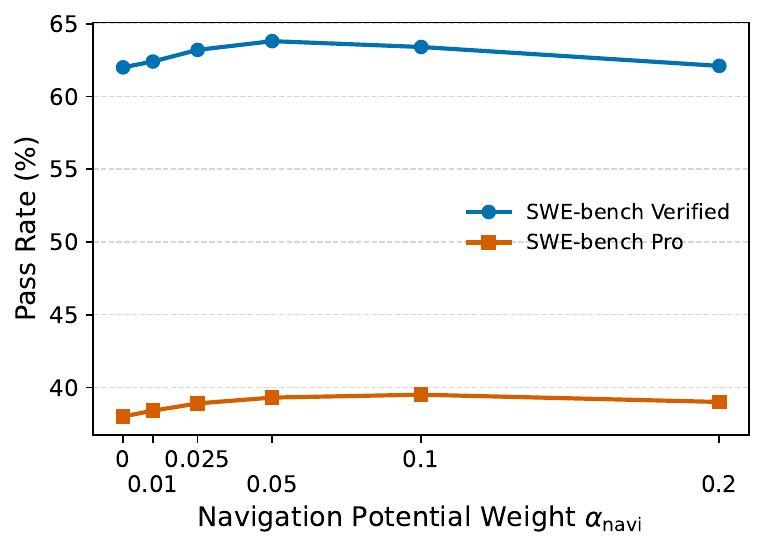}}
  \subfigure[Verification Potential Weight $\alpha_{\text{veri}}$]{\includegraphics[width=0.48\textwidth]{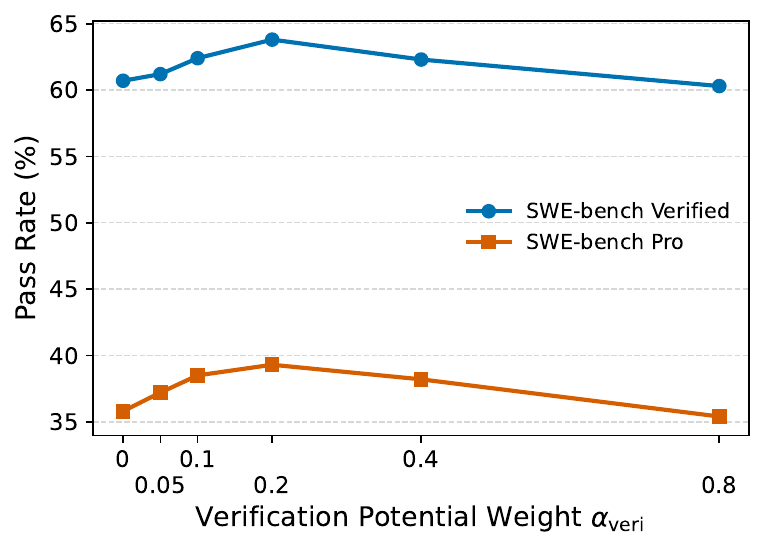}}
  \caption{Sensitivity to navigation and verification potential weights on SWE-bench Verified and SWE-bench Pro.}
  \label{fig:hyperparameter}
\end{figure*}

Figure~\ref{fig:hyperparameter} shows the effects of the navigation and verification potential weights on issue resolution. Performance on both benchmarks first improves and then declines as the weights increase, indicating that stronger process feedback is not always better and that overly strong potential signals may suppress exploration. Performance varies more smoothly over a moderate range of navigation weights, whereas excessively large verification weights cause more pronounced degradation, suggesting that although verification progress is an important signal, its reward scale requires tighter control.

\section{Conclusion}

We propose SWE-MILE, a process credit assignment method for reinforcement learning of long-horizon SWE agents. SWE-MILE characterizes task progress and functional regressions using navigation and verification potentials, and attributes feedback to specific actions through potential differences and backward credit, while asynchronous shadow probing allows intermediate verification to run in parallel with primary trajectory interaction. By retaining terminal outcome supervision without requiring an additional reward model or LLM evaluator, SWE-MILE provides a practical approach to obtaining fine-grained process supervision from SWE workflow runtime.

\subsection*{AI use statement}

We used generative AI tools to assist with parts of the software implementation, environment setup, and statistical analysis. AI tools were also used to polish the language and correct grammar in text drafted by the authors. The research idea, method, and experimental plan were developed by the authors without AI assistance. The authors reviewed all AI-assisted work and take full responsibility for the paper's code, analyses, claims, and text.




\subsection*{Reproducibility statement}

The Method section specifies the milestone potentials, process credit, policy objective, and asynchronous shadow-probing procedure. Appendix~\ref{sec:expdetailmain} reports the datasets, model settings, training hyperparameters, and evaluation budgets. Appendix~\ref{sec:expcodeflow} describes the Codeflow tools and attribution rules, while Appendix~\ref{sec:prompts} provides the agent prompts. The source code is publicly available. These details document the method and experimental protocol for reproduction.



\bibliography{iclr2027_conference}
\bibliographystyle{iclr2027_conference}

\appendix

\section*{Appendix Contents}

\begin{table}[!h]
    \centering
    \footnotesize
    \begin{tabular}{cl}
        \toprule
        \textbf{Appendix Sections} & \textbf{Contents} \\
        \midrule
        \autoref{sec:related_work} & Related Work \\
        \midrule
        \autoref{sec:expdetail} & Experimental Settings Details \\
        \midrule
        \autoref{sec:training_dynamics} & Training Dynamics \\
        \midrule
        \autoref{sec:discussion} & Further Discussions \\
        \midrule
        \autoref{sec:prompts} & Prompts and Instructions \\
        \bottomrule
    \end{tabular}
\end{table}

\section{Related Work}
\label{sec:related_work}

In this section, we review SWE benchmarks and verifiable environments, SWE agent scaffolds, and reinforcement learning and process reward methods for long-horizon agents.

\subsection{SWE Benchmarks and Scalable Verifiable Environments}

Recent research on coding agents has shifted from single-turn function-level code generation \citep{humaneval,mbpp,livecodebench} toward multi-turn repository-level issue resolution \citep{swebench,swebenchverified,swebenchpro} and even more challenging whole-repository generation \citep{nl2repobench,beyondswe,denovoswe} for SWE agents. Meanwhile, the construction of SWE agent datasets and benchmarks has evolved from static, manually curated evaluation sets toward scalable executable and verifiable training environments and automated data-generation pipelines, while gradually expanding from Python-centric ecosystems to multilingual software ecosystems. SWE-Gym \citep{swegym} first constructs real GitHub issues into repository-level training environments with executable runtimes and test-based verification, providing verifiable execution feedback for SWE agent post-training; R2E-Gym \citep{r2egym} further scales executable environments through procedural task generation and automated test generation; SWE-smith \citep{swesmith} automatically synthesizes software bugs in real codebases, scaling training tasks to tens of thousands. More recently, SWE-rebench V2 \citep{swerebenchv2} extends automated environment construction to 20 programming languages and thousands of repositories, while SWE-Universe \citep{sweuniverse} further demonstrates the feasibility of scaling real-world, verifiable multilingual SWE environments to the million scale. DeNovoSWE \citep{denovoswe} further extends scalable training environments to whole-repository generation tasks, providing large-scale training data for generating complete code repositories from natural-language specifications.

\subsection{SWE Agent Scaffolds and Environment Interaction}

Improvements in SWE agent capabilities have also benefited from the development of a series of agent scaffolds, which support long-horizon SWE task solving through structured tool use and environment interaction. Prior work has explored different agent-computer interfaces. SWE-agent \citep{sweagent} designs specialized interfaces for operations such as repository search, file viewing, code editing, and test execution; AutoCodeRover \citep{autocoderover} combines program structure with fault-localization information and provides structured retrieval interfaces over classes, methods, and code snippets to facilitate code localization. OpenHands \citep{openhands} further develops a composable and extensible platform to support SWE agents with rich tool-use capabilities. RepairAgent \citep{repairagent} designs specialized tools for program repair and uses a finite-state machine to constrain the set of tools available at different stages. In contrast, mini-SWE-agent \citep{sweagent} provides the model with only a general-purpose Bash interface, allowing it to compose shell commands for repository navigation, editing, and execution. Along a relatively independent design dimension, Agentless \citep{agentless} replaces open-ended agent–environment interaction with a predefined “localization–repair–validation” pipeline, showing that structured workflow decomposition can likewise achieve competitive performance.

\subsection{Credit Assignment for Long-Horizon SWE Agent Reinforcement Learning}

Reinforcement learning for long-horizon SWE agents typically uses the execution outcome of the final patch as a trajectory-level reward, making it difficult to identify effective steps in failed trajectories and redundant behaviors in successful ones. VeRPO \citep{verpo}, targeting function-level code generation tasks, converts per-test-case pass outcomes into dense verifiable rewards and mitigates cardinality bias through difficulty and density calibration. P2T \citep{p2t} uses a reference patch to reverse distill a process graph consisting of contextual facts and key repair progress, and employs multiple external LLM evaluators to offline filter candidate trajectories, thereby constructing privileged process-supervised SFT data. SWE-Shepherd \citep{sweshepherd} trains an action-level process reward model (PRM) using heuristic labels based on file access, code modifications, test feedback, and repetitive behavior, and uses the PRM to score candidate actions at inference time. SWE-TRACE \citep{swetrace} employs a Rubric Agent to generate issue-specific rubrics and external LLM judges to construct preference supervision for training a rubric-conditioned PRM, then combines its scores with terminal execution rewards for GRPO. Rubric-based GRM \citep{rubricgrm} directly uses Seed1.6 to evaluate candidate actions or trajectory segments based on human-written rubrics, the gold patch, and interaction history, and filters successful trajectories for RFT. G-RA \citep{gra} introduces rule-based immediate rewards for tool-call formatting and selection preferences when the terminal reward exceeds a threshold, mitigating policy degradation caused by reward misalignment, but does not characterize actual progress in repository state. CodePilot \citep{codepilot}, from a test-time search perspective, organizes partial patches as MCTS states and backpropagates values along the search tree based on execution signals such as syntactic validity, patch applicability, and test pass rate. More generally, GraphGPO \citep{graphgpo} aggregates multiple rollouts into a unified state-transition graph and assigns edge-level credit according to how much a transition reduces the graph distance to a successful state.

\section{Experimental Settings Details}
\label{sec:expdetail}

We summarize the main experimental settings. Unless otherwise specified, all experiments use the same hyperparameters for fair comparison.

\subsection{Main Setting}
\label{sec:expdetailmain}

We implement the agentic RL training pipeline of SWE-MILE based on the rLLM\footnote{\url{https://github.com/rllm-org/rllm}} \citep{rllm} framework, using the verl\footnote{\url{https://github.com/verl-project/verl}} \citep{verl} backend for policy optimization, vLLM\footnote{\url{https://github.com/vllm-project/vllm}} \citep{vllm} for rollout inference, and SWE-MiniSandbox to provide isolated code execution and asynchronous verification environments. Experiments are conducted on 8 compute nodes with a total of 64 GPUs (at least 80 GB device memory each). For the same task type, all compared methods use identical materialized training data, inference budgets, and general optimization settings, while ablation experiments modify only the corresponding components. We report the best performance among checkpoints evaluated every five training steps. The main training hyperparameters and evaluation settings are summarized in Table~\ref{tab:training_hyperparameters} and Table~\ref{tab:evaluation_settings}, respectively.

\begin{table}[!h]
    \centering
    \caption{Training hyperparameters for issue resolution and whole-repository generation.}
    \label{tab:training_hyperparameters}

    \small
    \setlength{\tabcolsep}{6pt}
    \renewcommand{\arraystretch}{1.0}
    \setlength{\heavyrulewidth}{0.8pt}
    \setlength{\lightrulewidth}{0.4pt}

    \begin{tabularx}{\linewidth}{
        @{}
        l
        *{2}{>{\centering\arraybackslash}X}
        @{}
    }
        \toprule
        \textbf{Hyperparameter}
        & \makecell{\textbf{Issue}\\\textbf{Resolution}}
        & \makecell{\textbf{Whole-Repository}\\\textbf{Generation}} \\
        \midrule

        \multicolumn{3}{c}{\textit{Policy Optimization}} \\
        \midrule

        Initial policy model             & Qwen3.5-9B & Qwen3.5-35B-A3B \\
        Optimizer                        & Adam & Adam \\
        Learning rate                    & 1e-6 & 1e-6 \\
        Learning rate schedule           & Constant & Constant \\
        Training Dataset                 & SWE-rebench V2 & DeNovoSWE \\
        Training batch size (tasks)      & 12 & 24 \\
        Optimization epochs              & 1 & 1 \\
        PPO clipping $(\epsilon_{-}, \epsilon_{+})$  & $(0.2, 0.28)$ & $(0.2, 0.28)$ \\
        KL penalty coefficient           & 0.001 & 0.001 \\
        Entropy coefficient              & 0 & 0 \\
        Gradient clipping norm           & 1.0 & 1.0 \\

        \midrule
        \multicolumn{3}{c}{\textit{Training Rollouts}} \\
        \midrule

        Rollouts per task ($K$)          & 16 & 16 \\
        Sampling temperature             & 1.0 & 1.0 \\
        Top-$p$                          & 0.95 & 0.95 \\
        Top-$k$                          & 20 & 20 \\
        Maximum turns                    & 200 & 500 \\
        Maximum context length (tokens)  & 256k & 256k \\
        Maximum output tokens per turn   & 8192 & 16384 \\
        Agent scaffold                   & Codeflow & Codeflow \\

        \midrule
        \multicolumn{3}{c}{\textit{Milestone Reward}} \\
        \midrule

        Navigation path exposure score ($q^{\text{observed}}$) & 0.2 & -- \\
        Navigation content exposure score ($q^{\text{observed}}$) & 1.0 & -- \\
        Any P2P regression penalty ($\beta_{\text{any}}$) & 0.1 & -- \\
        P2P regression fraction penalty ($\beta_{\text{frac}}$) & 0.5 & -- \\
        Navigation potential weight ($\alpha_{\text{navi}}$) & 0.05 & Disabled \\
        Verification potential weight ($\alpha_{\text{veri}}$) & 0.2 & 0.1 \\
        Potential difference clip ($-c_{-},c_{+}$) & -- & $(-6,3)$ \\
        Backward credit weight ($\lambda$) & 0.2 & 0.2 \\
        Backward credit discount ($\gamma$) & 0.9 & 0.9 \\
        Output-format reward weight ($\eta$) & 0.25 & 0.25 \\
        Invalid output-format reward ($r_t^{\text{fmt}}$) & $-1$ & $-1$ \\

        \bottomrule
    \end{tabularx}
\end{table}

\begin{table}[!h]
    \centering
    \caption{Evaluation settings for issue resolution
    and whole-repository generation.}
    \label{tab:evaluation_settings}

    \small
    \setlength{\tabcolsep}{6pt}
    \renewcommand{\arraystretch}{1.0}
    \setlength{\heavyrulewidth}{0.8pt}
    \setlength{\lightrulewidth}{0.4pt}

    \begin{tabularx}{\linewidth}{
        @{}
        l
        *{2}{>{\centering\arraybackslash}X}
        @{}
    }
        \toprule
        \textbf{Parameter}
        & \makecell{\textbf{Issue}\\\textbf{Resolution}}
        & \makecell{\textbf{Whole-Repository}\\\textbf{Generation}} \\
        \midrule

        Sampling temperature             & 0.6 & 0.6 \\
        Top-$p$                          & 0.95 & 0.95 \\
        Top-$k$                          & 20 & 20 \\
        Maximum turns                    & 200 & 500 \\
        Maximum context length (tokens)  & 256k & 256k \\
        Maximum output tokens per turn   & 8192 & 16384 \\
        Budget per task                  & 4 & 4 \\
        Agent scaffold                   & Codeflow & Codeflow \\

        \bottomrule
    \end{tabularx}
\end{table}

\noindent\textbf{Dynamic Sampling.} During training, we adopt an \textit{outcome-driven asymmetric dynamic sampling strategy} to improve data utilization. Specifically, we generate a group of rollouts for each task and filter task groups according to verifier outcomes before advantage computation. For issue resolution, if all trajectories in a group receive terminal rewards of either 0 or 1, the group lacks an outcome-level comparison signal and the task is filtered and returned to the sampling pool. For whole-repository generation, the same filtering is applied if all trajectories have the same number of passed verifier tests, or if their test pass rates are all close to either full or zero. Filtered tasks are resampled for a limited number of attempts, with easy tasks stopping earlier and difficult tasks receiving more retry opportunities. For example, easy tasks receive only one sampling attempt, while difficult tasks receive two. The remaining task groups are used to form the training batch, while asynchronous sampling continuously adds candidate tasks until the required batch size is reached. This strategy makes difficult tasks more likely to be resampled by an updated policy at middle and later training steps, giving them more opportunities to produce successful trajectories as the policy improves.

\noindent\textbf{Reward at Turn or Context Limits.} For issue resolution trajectories that hit the turn or context limit but still pass the verifier, we discount the terminal outcome reward to account for potentially redundant steps after the verification target has been reached. Let $t^*$ denote the first turn at which the verification potential reaches 1 from below 1. We set the reward to $R^{\mathrm{lim}}=t^*/|\mathcal{T}|$. The earlier the target is reached and the more subsequent steps remain, the lower the reward. In the outcome only ablation training, this reward is uniformly set to $0.6$.

\subsection{Baseline Implementation}
\label{sec:baselineimp}

For a fair comparison, we reimplement all baseline methods under the same experimental conditions as SWE-MILE. We next describe the key implementation details of each baseline.

\noindent\textbf{GRPO} We augment GRPO based on outcome rewards with the same format reward adjustment to the advantage as in SWE-MILE to stabilize long-horizon agentic training.

\noindent\textbf{G-RA} We replace the group-normalized advantage estimation in G-RA with RLOO-style advantage estimation to align it with SWE-MILE, thereby isolating the process reward design as the primary variable. Since G-RA does not provide open-source code, we implement the method according to the design described in its paper.

\noindent\textbf{SWE-TRACE} The training of the rubric generator and trajectory judge in SWE-TRACE essentially distills MiniMax M2.5. Since our experiments focus on comparing rubric-based reward methods with SWE-MILE, we directly use an open-source model stronger than MiniMax M2.5 (namely Qwen3.8-27B \citep{qwen3.8}) as the rubric generator and trajectory judge, rather than training them separately. In addition, SWE-TRACE does not provide accessible open-source code, so we implement the method following the procedure described in its paper.

\noindent\textbf{GiGPO and GraphGPO} SWE tasks have a rich state space, with states that are high-dimensional, history-dependent, continuously changing, and partially observable, making state matching more difficult than in environments such as ALFWorld \citep{alfworld} and WebShop \citep{webshop}. To adapt state aggregation methods such as GiGPO and GraphGPO to SWE tasks, we use a more compact state definition. We define the state by the numbers of passing F2P and P2P tests and the exposure states of relevant files through search and inspection. For tests, we consider only the number of passing F2P tests and the number of passing P2P tests separately, rather than requiring exact per-test state matching or combining the two counts. Search exposure and inspection exposure of relevant files are treated as distinct. Specifically, the state at step $t$, denoted by $z_t$, is defined as:
\begin{equation}
z_t= \left( N^{\text{F2P}}_t,\; N^{\text{P2P}}_t,\; \mathcal{S}_t,\; \mathcal{V}_t \right),
\end{equation}
where $N^{\text{F2P}}_t$ is the number of passing F2P tests for the current task, $N^{\text{P2P}}_t$ is the number of passing P2P tests, $\mathcal{S}_t$ is the set of relevant files exposed to the agent through search, and $\mathcal{V}_t$ is the set of relevant files actually inspected by the agent. For whole-repository generation, file-exposure states are omitted. This design avoids an overly strict state definition that would make matching equivalent steps difficult. These test states are also obtained using the shadow probing mechanism described in Section~\ref{sec:shadow_probe}. Therefore, we instantiate shadow sandboxes for GiGPO and GraphGPO during training. 

All baseline methods use the same materialized datasets and experimental configurations as SWE-MILE to compare the core method designs while keeping other conditions as consistent as possible.

\subsection{Codeflow Scaffold}
\label{sec:expcodeflow}

Codeflow organizes repository interaction into tool calls with explicit arguments, and uniformly records the tool type, execution result, file exposure evidence, and repository state changes as action events to support subsequent milestone attribution. Codeflow supports two configurations: structured full-tool and bash-only. The former provides all tools listed in Table~\ref{tab:codeflow_tools}; the latter retains only \texttt{execute\_bash} as the environment interaction tool and \texttt{submit} as the termination interface.

\begin{table}[!h]
    \centering
    \caption{Codeflow tools and modes. ``--'' indicates that the tool has no mode parameter.}
    \label{tab:codeflow_tools}

    \small
    \setlength{\tabcolsep}{6pt}
    \renewcommand{\arraystretch}{1.0}
    \setlength{\heavyrulewidth}{0.8pt}
    \setlength{\lightrulewidth}{0.4pt}

    \begin{tabularx}{\linewidth}{
        @{}
        l
        l
        >{\raggedright\arraybackslash}X
        @{}
    }
        \toprule
        \textbf{Tool} & \textbf{Mode} & \textbf{Description} \\
        \midrule

        \multirow{4}{*}{\texttt{search}}
        & \texttt{text} & Find case-sensitive literal text with surrounding context. \\
        & \texttt{symbol} & Match symbols at identifier boundaries with surrounding context. \\
        & \texttt{file} & Find file paths by filename substring. \\
        & \texttt{directory} & Find directory paths by directory name substring. \\

        \midrule
        \multirow{3}{*}{\texttt{read\_file}}
        & \texttt{file} & Read a line range with line numbers. \\
        & \texttt{diff} & Show the diff against Git HEAD, optionally scoped to a path. \\
        & \texttt{status} & Show Git status, including untracked files, optionally scoped to a path. \\

        \midrule
        \multirow{6}{*}{\texttt{edit\_file}}
        & \texttt{replace} & Replace exact text after checking the expected match count. \\
        & \texttt{insert} & Insert text after a given line. \\
        & \texttt{create} & Create a new file with the supplied content. \\
        & \texttt{delete} & Delete the specified file. \\
        & \texttt{apply\_patch} & Atomically apply a Git-compatible patch across one or more files. \\
        & \texttt{undo} & Undo the most recent successful \texttt{edit\_file} action. \\

        \midrule
        \texttt{execute\_bash} & -- & Run tests and Shell commands, subject to configured file-operation restrictions. \\
        \texttt{submit} & -- & Submit the current repository state and end the interaction. \\

        \bottomrule
    \end{tabularx}
\end{table}

\noindent\textbf{Search and Read Attribution.} For dedicated tools, attribution is based on the observation ultimately presented to the model after successful execution. Matching paths and snippets returned by \texttt{search} are treated as search exposure, while file contents or diff hunks returned by \texttt{read\_file} are treated as content exposure. Outputs containing only file paths or status are not considered actual content inspection. For Bash, Codeflow uses lexical tokenization based on \texttt{shlex} and command rules to identify search and read operations, and applies Python AST analysis to inline Python code. It then performs file-level attribution by combining the model visible stdout, parsed target paths, and a trusted repository file inventory. Only high-confidence evidence is used to update navigation potential. Path exposure and content exposure are distinguished during updates, and the maximum potential attained by each relevant file is retained, so repeated exposure does not further increase the potential.

\noindent\textbf{Edit Attribution.} Codeflow compares repository snapshot fingerprints before and after tool execution and records the changed files, attributing actual repository state changes to the current action. This allows file changes caused by Bash to be identified without relying on explicit editing syntax in the command. Both tool configurations prohibit modifications to protected test files or changes to Git HEAD to preserve consistency between the evaluation basis and the initial code version. Violating actions are rejected and reverted and penalized through the format reward.

\subsection{Shadow Probe for Whole-Repository Generation}
\label{sec:denovo_shadow_probe}

\begin{wraptable}{r}{0.56\textwidth}
    \vspace{-2em}
    \centering
    \caption{Runtime statistics of probe merge and multi-shadow probing for whole-repository generation.}
    \label{tab:denovo_shadow_runtime}
    \small
    \renewcommand{\arraystretch}{1.05}
    \setlength{\tabcolsep}{4pt}
    \setlength{\heavyrulewidth}{0.8pt}
    \setlength{\lightrulewidth}{0.4pt}

    \resizebox{\linewidth}{!}{
    \begin{tabular}{lrrr}
        \toprule
        \textbf{Runtime Component} & \textbf{Mean} & \textbf{Unit} & \textbf{Samples} \\
        \midrule
        Agent generation - turn & 6.30 s & turn & 314,547 \\
        Agent generation - traj & 704.84 s & rollout & 2,814 \\
        Action execution & 5.30 s & turn & 314,547 \\
        Primary agent interaction & 1,761.83 s & rollout & 2,814 \\
        \hdashline
        Shadow verifier probe & 308.77 s & probe & 74,673 \\
        Primary final verifier & 298.90 s & rollout & 2,814 \\
        \hdashline
        Probe saved by merge & 2.94 & rollout & 2,814 \\
        Shadow drained at primary finish & 35.11\% & rollout & 2,814 \\
        Shadow drained at batch finish & 96.48\% & rollout & 2,814 \\
        \bottomrule
    \end{tabular}
    }
\end{wraptable}

Whole-repository generation tasks require progressively building complete functionality from an initial repository and therefore involve more editing steps than issue resolution tasks, triggering more frequent shadow probes. Their heavier verifiers typically require longer execution time. To improve verification efficiency and shadow probe concurrency, we combine \textit{probe merge} and \textit{multi-shadow parallel probing} during DeNovoSWE training, and defer the final synchronization of shadow probes until the batch rollout is completed. Specifically, for consecutive repository modification actions that satisfy the replay conditions, if their non-empty changed file sets are identical, we merge them into a single probe group. Actions within the group are still replayed in their original order, but the verifier is executed only once on the repository state corresponding to the final action, and the resulting net potential change is evenly assigned to all actions in the group. Meanwhile, we configure multiple independent shadow sandboxes for each rollout, which replay to the repository states corresponding to different probes and execute the verifier in parallel. Results from all shadows are finally aggregated in the original action order to preserve the temporal consistency between potential changes and credit assignment. Shadow probes may continue running after the current rollout finishes, overlapping with the execution of other rollouts in the same rollout batch, and are synchronized before the policy update. Table~\ref{tab:denovo_shadow_runtime} reports runtime statistics for DeNovoSWE training with probe merge and four parallel shadow sandboxes, including verifier runtime, the number of probes saved, and the achieved concurrency.

\section{Training Dynamics}
\label{sec:training_dynamics}

In this section, we further examine evaluation performance on SWE-bench Pro benchmark, interaction behavior during training rollouts, and the policy optimization process, providing three perspectives on model performance, interaction cost, and optimization dynamics.

\subsection{Evaluation Dynamics on SWE-bench Pro}

\begin{figure*}[!h]
  \centering
  \subfigure[Pass Rate]{\includegraphics[width=0.245\textwidth]{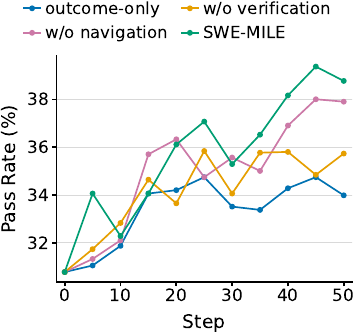}}
  \subfigure[Turns]{\includegraphics[width=0.245\textwidth]{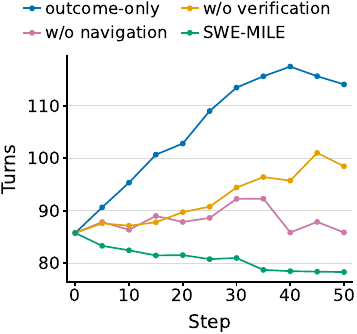}}
  \subfigure[Response Length]{\includegraphics[width=0.245\textwidth]{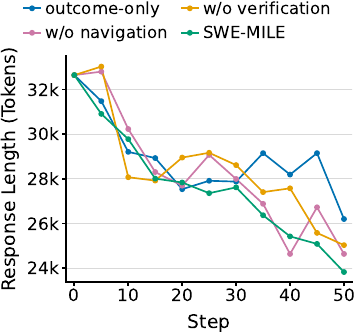}}
  \subfigure[Trajectory Length]{\includegraphics[width=0.245\textwidth]{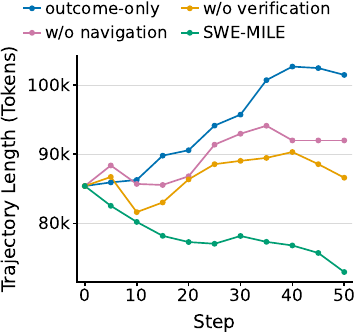}}
  \caption{SWE-bench Pro evaluation dynamics across training checkpoints.}
  \label{fig:swebench_pro_eval_dynamics}
\end{figure*}

Figure~\ref{fig:swebench_pro_eval_dynamics} compares outcome-only, the two potential ablations, and SWE-MILE at checkpoints every 5 steps in terms of average pass rate, turns, output tokens, and total trajectory tokens. SWE-MILE shows an overall increase in success rate while interaction turns and token counts gradually decrease. Outcome-only training also improves in success rate, but its interaction turns continue to increase, and total trajectory length still grows despite fewer model output tokens, indicating a larger proportion of observation tokens. Removing either potential also results in less pronounced trajectory shortening than the full method. Overall, these trends are consistent with the design goal of milestone credit: helping the policy complete repairs more directly while reducing redundant interactions.

\subsection{Training Rollout Dynamics}

\begin{figure*}[!h]
  \centering
  \subfigure[Turns]{\includegraphics[width=0.49\textwidth]{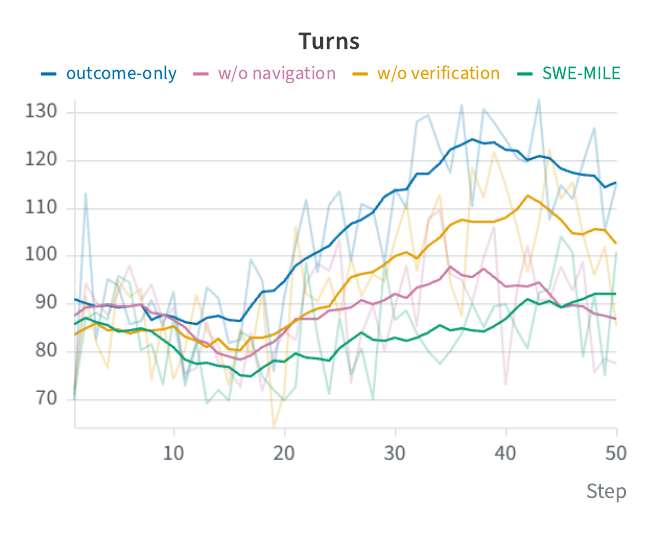}}%
  \hspace{0.005\textwidth}%
  \subfigure[Within-Budget Rollout Ratio]{\includegraphics[width=0.49\textwidth]{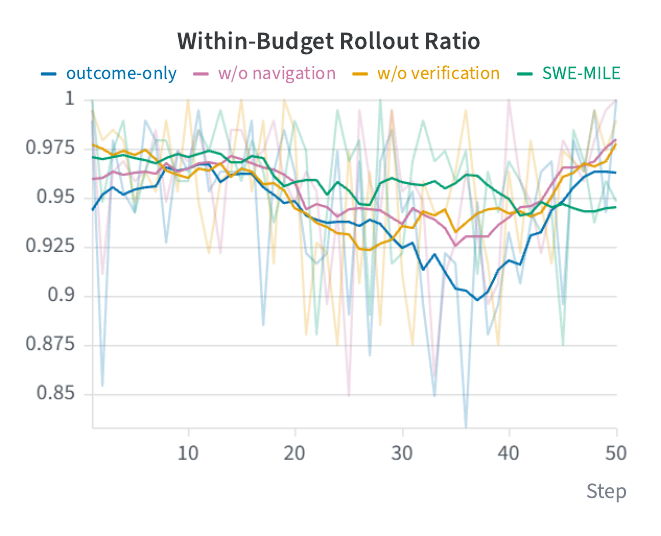}}
  \par\vspace{-1mm}
  \subfigure[Response Length (Tokens)]{\includegraphics[width=0.49\textwidth]{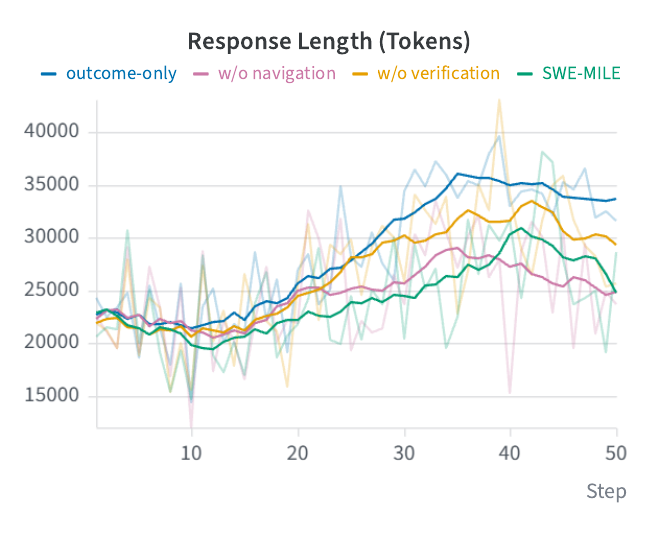}}%
  \hspace{0.005\textwidth}%
  \subfigure[Trajectory Length (Tokens)]{\includegraphics[width=0.49\textwidth]{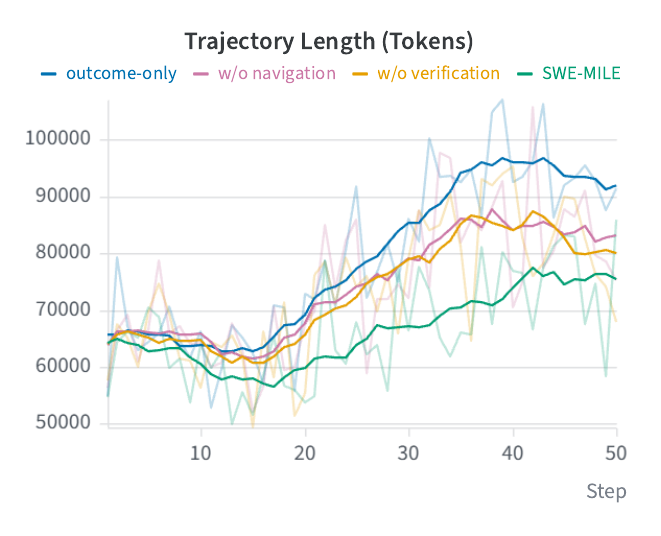}}
  \caption{Training rollout dynamics across reward settings.}
  \label{fig:training_rollout_dynamics}
\end{figure*}

Figure~\ref{fig:training_rollout_dynamics} shows how interaction turns, the proportion of rollouts within the inference budget, model output length, and total trajectory length evolve during training under four training settings. In the middle and later stages, trajectories grow under all methods, mainly because the asymmetric dynamic sampling strategy gives difficult tasks more retry opportunities. As a result, the larger share of difficult tasks later in training requires more interaction turns and output tokens. However, outcome-only shows the largest increase in interaction turns and trajectory length, while the proportion of rollouts within the budget also drops at one point. SWE-MILE exhibits slower trajectory growth. Removing either potential generally increases interaction cost relative to the full method, with a particularly noticeable increase in turns when verification potential is removed. These observations indicate that milestone credit helps suppress trajectory expansion during training, consistent with the SWE-bench Pro evaluation results in the previous subsection.

\subsection{Policy Optimization Dynamics}

\begin{figure*}[!h]
  \centering
  \subfigure[Loss]{\includegraphics[width=0.49\textwidth]{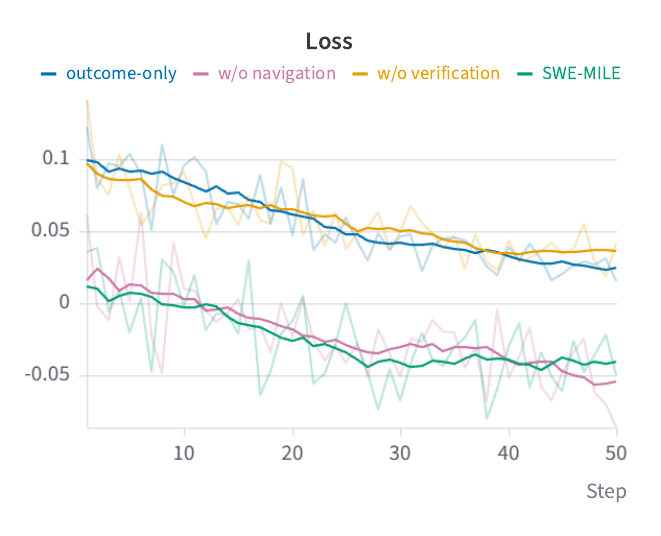}}%
  \hspace{0.005\textwidth}%
  \subfigure[Gradient Norm]{\includegraphics[width=0.49\textwidth]{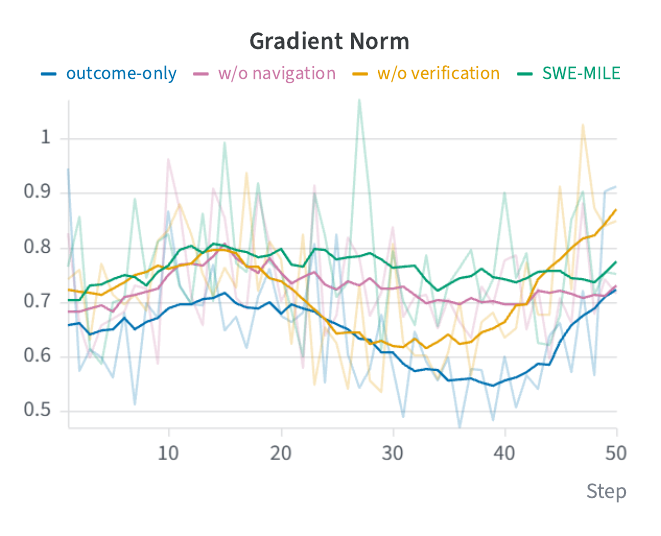}}
  \par\vspace{-1mm}
  \subfigure[KL Divergence]{\includegraphics[width=0.49\textwidth]{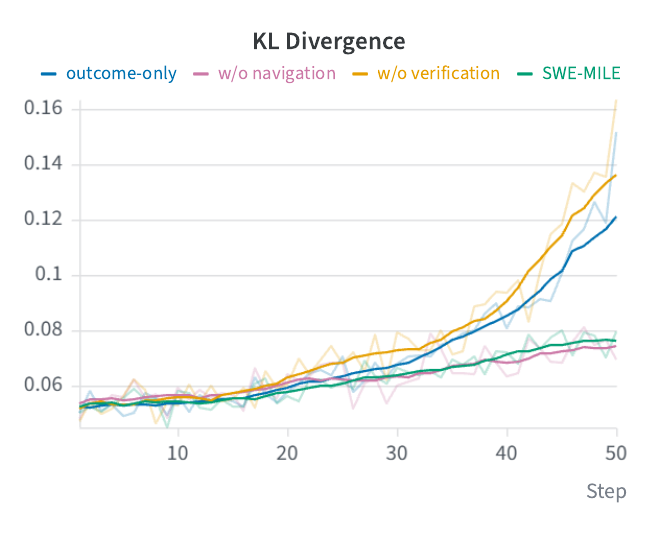}}%
  \hspace{0.005\textwidth}%
  \subfigure[Policy Entropy]{\includegraphics[width=0.49\textwidth]{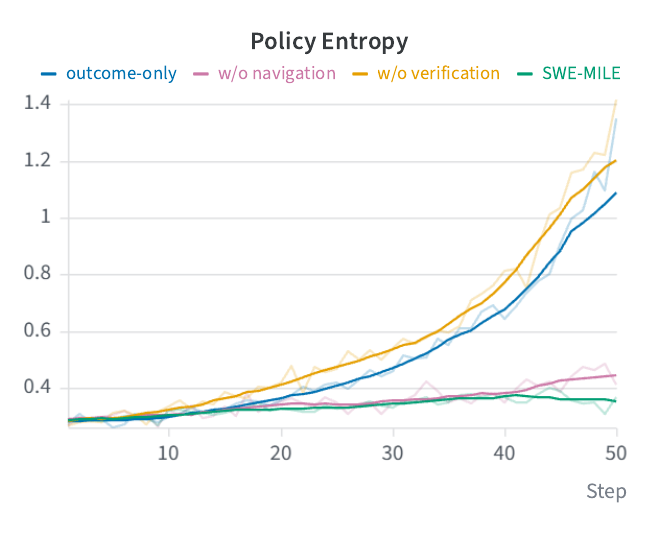}}
  \caption{Policy optimization dynamics across reward settings.}
  \label{fig:policy_optimization_dynamics}
\end{figure*}

Figure~\ref{fig:policy_optimization_dynamics} compares policy loss, gradient norm, KL divergence, and policy entropy under four training settings. The gradient norms remain at a similar scale across methods, while KL and entropy exhibit different trends. Both outcome-only and w/o verification increase noticeably in the later stages of training, whereas SWE-MILE and w/o navigation remain relatively stable. The latter two settings both retain verification potential, suggesting that intermediate feedback from test states may help constrain policy changes, consistent with their slower trajectory growth in the previous subsection. Since the methods use different advantage signals, their loss values show clear separation, and the absolute loss values should not be directly compared as a measure of performance.

\section{Discussion}
\label{sec:discussion}

This section discusses several additional considerations related to SWE-MILE.

\subsection{Replay Fidelity of SWE Environments}

A practical property that makes asynchronous shadow probing particularly suitable for SWE agents is the high replay fidelity of SWE environments. Unlike web or GUI environments, where state transitions may be affected by transient interface elements, remote services, or other external events \citep{web1,gui1,gui2}, SWE interactions primarily modify repository and execution states inside controlled sandboxes. Given the same initial repository state, environment configuration, and sequence of repository-changing actions, intermediate repository states can therefore typically be reconstructed with high fidelity. This enables SWE-MILE to decouple verification from the primary rollout: repository-changing actions can be replayed asynchronously in a shadow sandbox, where the verifier evaluates the reconstructed intermediate states without blocking agent interaction. More generally, the effectiveness of asynchronous shadow probing depends on the reproducibility of environment transitions, making it particularly well suited to sandboxed SWE environments, while its direct extension to more stochastic agent environments may require explicit state snapshotting, replay validation, or mechanisms for handling environmental divergence.

\subsection{Limitations}

SWE-MILE has the following limitations:
\begin{itemize}
\item \textbf{Availability of fine-grained verification feedback.} Test frameworks across different programming languages produce heterogeneous result formats, and each language typically requires a dedicated verification potential collection pipeline to be developed and validated.
\item \textbf{Differences in test importance.} The current verification potential aggregates results within each test category by pass rate or count, without distinguishing the importance of individual tests to task completion. As a result, different test states may receive the same potential value even when they represent different levels of functional progress.
\end{itemize}

\section{Prompts}
\label{sec:prompts}

In this section, we present the prompts used in our experiments. Placeholders substituted at runtime are shown in \textcolor{varblue}{blue}.

\subsection{System Prompt}
\label{sec:prompt_system}

The system prompt declares the agent role, constrains the interaction protocol to exactly one native tool call per assistant turn, instantiates the tool policy, specifies the path convention, and explains the semantics of tool observations.

\begin{promptbox}{Issue Resolution System Prompt}
\texttt{You are a programming agent responsible for resolving an issue in the repository at \textcolor{varblue}{repository\_root}.\\
\\
Interact with the repository only through the provided native tools. Call exactly one tool per assistant turn and never write a tool call as text.\\
\\
Tool policy:\\
- Prefer the dedicated search, read\_file, and edit\_file tools for repository exploration, reading, and editing.\\
- Use search as the default for repository searches and directory exploration. Avoid grep, rg, find, fd, git grep, ls/tree, or equivalent shell/Python searches when search can perform the operation. Bash search is a fallback only when the dedicated tool cannot express the required operation.\\
- Use read\_file as the default for repository file content, diffs, and status. Avoid cat, head/tail, sed/awk, Python file reads, git diff/status/show, or equivalent commands when read\_file can perform the operation. Bash reading is a fallback only when the dedicated tool cannot express the required operation.\\
- Use edit\_file as the default for creating, modifying, and deleting repository files. Avoid shell redirection, sed -i, Python file writes, or equivalent commands when edit\_file can perform the operation. Bash editing is a fallback only when the dedicated tool cannot express the required operation.\\
- Use execute\_bash primarily for tests, builds, and commands not covered by the dedicated tools. Put temporary scripts and outputs under /tmp. Searches and reads outside the repository and filtering command output remain allowed. Follow the protected-test, Git HEAD, environment, download, timeout, and sandbox boundary rules.\\
- Use submit only after the implementation has been verified. Call it exactly as submit(\{\}) with no arguments; do not add summary, answer, result, or other fields.\\
\\
Paths may be relative to \textcolor{varblue}{repository\_root} or absolute paths inside it. Read relevant code before editing, make the smallest focused change that solves the task, inspect the resulting diff, run relevant tests, and then submit.\\
\\
Search and file reads are paginated. If relevant information remains, continue search with next\_offset as offset, or continue a file read with next\_start\_line as start\_line. A truncated result can also mean that a very long line or the middle of a large output was clipped; omission markers report how much was omitted.\\
\\
The default execute\_bash command timeout is \textcolor{varblue}{command\_timeout} seconds.}
\end{promptbox}

\begin{promptbox}{Whole-Repository Generation System Prompt}
\texttt{You are a programming agent responsible for implementing a complete software package from the supplied specification in the workspace at \textcolor{varblue}{repository\_root}.\\
\\
This is a repository-generation task. The workspace may contain starter files or an incomplete implementation. Build all functionality required by the specification, rather than assuming there is an existing implementation with a single bug to fix.\\
\\
Interact with the repository only through the provided native tools. Call exactly one tool per assistant turn and never write a tool call as text.\\
\\
Tool policy:\\
\textcolor{varblue}{[the same instantiated tool policy as in the issue resolution system prompt]}\\
\\
Paths may be relative to \textcolor{varblue}{repository\_root} or absolute paths inside it. Read the specification and inspect the workspace first. Identify the required public interfaces, behaviors, edge cases, and packaging requirements; implement the necessary modules and files as a coherent package. Use existing starter files when appropriate, but do not limit the work to a small patch when the specification requires broader implementation.\\
\\
Verify the implementation against the specification with relevant tests and checks, inspect the final files and diff, and then submit the repository state. Preserve protected acceptance tests; do not alter them or substitute hard-coded test answers for the required functionality.\\
\\
Search and file reads are paginated. If relevant information remains, continue search with next\_offset as offset, or continue a file read with next\_start\_line as start\_line. A truncated result can also mean that a very long line or the middle of a large output was clipped; omission markers report how much was omitted.\\
\\
The default execute\_bash command timeout is \textcolor{varblue}{command\_timeout} seconds.\\
\\
Repository-generation source policy:\\
- Implement the target package only from the supplied specification.\\
- Do not install, download, clone, inspect, or extract the target package's upstream source. Third-party dependencies and a local editable install of your own implementation are allowed.}
\end{promptbox}

\subsection{Issue Resolution Prompt}
\label{sec:prompt_issue}

For issue resolution, the user prompt supplies the repository root and the natural-language issue description of the task instance, and instructs the agent to inspect and update the repository with the Codeflow tools, verify the result with relevant tests, and submit the final repository state. The issue description is used verbatim as materialized by the dataset, without adding fault localization hints, reproduction scripts, or references to the ground-truth patch or the hidden tests.

\begin{promptbox}{Issue Resolution Prompt}
\texttt{The repository root is:\\
\textcolor{varblue}{repository\_root}\\
\\
Resolve the following software-engineering task:\\
\\
<issue\_description>\\
\textcolor{varblue}{problem\_statement}\\
</issue\_description>\\
\\
Use the available codeflow tools to inspect and update the repository, verify the result with relevant tests, and submit the final repository state.}
\end{promptbox}

\subsection{Whole-Repository Generation Prompt}
\label{sec:prompt_repogen}

For whole-repository generation, the user prompt supplies the workspace root and the package specification, and asks the agent to create or complete the repository files required by the specified interfaces and behavior. Compared with issue resolution, the instruction emphasizes constructing functionality from the specification rather than locating and repairing a defect, which is consistent with the longer horizon and the larger number of repository-changing actions observed in this task family.

\begin{promptbox}{Whole-Repository Generation Prompt}
\texttt{The implementation workspace is:\\
\textcolor{varblue}{repository\_root}\\
\\
Implement the software package described by the following specification:\\
\\
<package\_specification>\\
\textcolor{varblue}{problem\_statement}\\
</package\_specification>\\
\\
Create or complete the repository files needed to satisfy the specified interfaces and behavior. Use the available codeflow tools to inspect the workspace, implement the package, verify it with relevant tests, and submit the final repository state.}
\end{promptbox}

\end{document}